\documentclass{bmvc2k}

\title{Graph Based Convolutional Neural Network}

\addauthor{Michael Edwards}{}{1}
\addauthor{Xianghua Xie}{http://www.csvision.swan.ac.uk}{1}

\addinstitution{
 Swansea University\\
 Swansea, UK
 }

\runninghead{Edwards, Xie}{Graph Convolutional Neural Network}

\usepackage{times}
\usepackage{graphicx}
\usepackage{epsfig}
\usepackage{epstopdf}
\usepackage{amsfonts}
\usepackage{multirow}
\usepackage{hhline}
\usepackage[nolist, nohyperlinks]{acronym}
\usepackage{colortbl}
\definecolor{Gray}{gray}{0.85}
\usepackage{rotating}
\usepackage{fix2col}
\usepackage{pdflscape}
\usepackage{array}
\usepackage{tabu}
\usepackage{longtable}
\usepackage{pdfpages}
\usepackage[export]{adjustbox}
\usepackage{blindtext}
\usepackage{algorithm,algorithmic}
\renewcommand{\mathbf }{\boldsymbol}

\begin{document}

\maketitle

\begin{abstract}
The benefit of localized features within the regular domain has given rise to the use of \acp{CNN} in machine learning, with great proficiency in the image classification. The use of \acp{CNN} becomes problematic within the irregular spatial domain due to design and convolution of a kernel filter being non-trivial. One solution to this problem is to utilize graph signal processing techniques and the convolution theorem to perform convolutions on the graph of the irregular domain to obtain feature map responses to learnt filters. We propose graph convolution and pooling operators analogous to those in the regular domain. We also provide gradient calculations on the input data and spectral filters, which allow for the deep learning of an irregular spatial domain problem. Signal filters take the form of spectral multipliers, applying convolution in the graph spectral domain. Applying smooth multipliers results in localized convolutions in the spatial domain, with smoother multipliers providing sharper feature maps. Algebraic Multigrid is presented as a graph pooling method, reducing the resolution of the graph through agglomeration of nodes between layers of the network. Evaluation of performance on the MNIST digit classification problem in both the regular and irregular domain is presented, with comparison drawn to standard \ac{CNN}. The proposed graph \ac{CNN} provides a deep learning method for the irregular domains present in the machine learning community, obtaining 94.23\% on the regular grid, and 94.96\% on a spatially irregular subsampled MNIST. 
\end{abstract}

\begin{acronym}
\acro{HAR}{Human Action Recognition}
\acro{SoKP}{Sequence of Key Poses}
\acro{BoKP}{Bag of Key Poses}
\acro{DTW}{Dynamic Time Warping}
\acro{DMW}{Dynamic Manifold Warping}
\acro{CTW}{Canonical Time Warping}
\acro{ACA}{Aligned Cluster Analysis}
\acro{HACA}{Hierarchical Aligned Cluster Analysis}
\acro{EA}{Evolutionary Algorithm}
\acro{TPID}{Two Person Interaction Dataset}
\end{acronym}
\begin{acronym}
\acro{AMG}{Algebraic Multigrid}
\acro{CNN}{Convolutional Neural Network}
\acro{GCNN}{Graph-based Convolutional Neural Network}
\acrodefplural{CNN}[CNNs]{Convolutional Neural Networks}
\acro{GFT}{Graph Fourier Transform}
\acro{LOAO}{Leave One Actor Out}
\acro{LOSO}{Leave One Sequence Out}
\acro{SoKP}{Sequence of Key Poses}
\acro{BoKP}{Bag of Key Poses}
\acro{DTW}{Dynamic Time Warping}
\acro{DMW}{Dynamic Manifold Warping}
\acro{CTW}{Canonical Time Warping}
\acro{ACA}{Aligned Cluster Analysis}
\acro{HACA}{Hierarchical Aligned Cluster Analysis}
\acro{EA}{Evolutionary Algorithm}
\acro{MoCap}{Motion Capture}
\acro{IMU}{Inertial Measurement Unit}
\acro{HOG}{Histogram of Oriented Gradients}
\acro{HOF}{Histograms of Optical Flow}
\acro{STIP}{Space-Time Interest Point}
\acro{SVM}{Support Vector Machine}
\acro{SIFT}{Scale Invariant Feature Transform}
\acro{SotA}{State-of-the-Art}

\acro{HAR}{Human Action Recognition}

\acro{CMU}{Carnegie Mellon University}
\acro{TUM}{Technische Universit\"{a}t M\"{u}nchen}
\acro{MSR}{Microsoft Research}
\acro{BIT}{Beijing Institute of Technology}
\acro{INRIA}{Institut National de Recherche en Informatique et en Automatique}
\acro{JPL}{Jet Propulsion Laboratory}

\acro{TPID}{Two Person Interaction Dataset}
\acro{CAVIAR}{Context Aware Vision using Image-based Active Recognition}
\acro{IXMAS}{INRIA Xmas Motion Acquisition Sequences}
\acro{MuHAVi}{Multicamera Human Action Video}
\acro{AoDL}{Activities of Daily Living}
\acro{UMPM}{Utrecht Multi-Person Benchmark}
\acro{MHAD}{Multimodal Human Action Database}

\end{acronym}
\section{Introduction}In recent years, the machine learning and pattern recognition community has seen a resurgence in the use of neural network and deep learning architecture for the understanding of classification problems. Standard fully connected neural networks have been utilized for domain problems within the feature space with great effect, from text document analysis to genome characterization \cite{Zhang2006NNGenomes}. The introduction of the \ac{CNN} provided a method for identifying locally aggregated features by utilizing kernel filter convolutions across the spatial dimensions of the input to extract feature maps \cite{Lecun1998CNN}. Applications of \acp{CNN} have shown strong levels of recognition in problems from face detection \cite{Li2015CNNCascade}, digit classification \cite{Ciresan2012MCDNN}, and classification on a large number of classes \cite{Oquab2014CNNTransfer}.

The core \ac{CNN} concept introduces the hidden convolution and pooling layers to identify spatially localized features via a set of receptive fields in kernel form. The convolution operator takes an input and convolves kernel filters across the spatial domain of the data provided some stride and padding parameters, returning feature maps that represent response to the filters. Given a multi-channel input, a feature map is the summation of the convolutions with separate kernels for each input channel. In \ac{CNN} architecture, the pooling operator is utilized to compress the resolution of each feature map in the spatial dimensions, leaving the number of feature maps unchanged. Applying a pooling operator across a feature map enables the algorithm to handle a growing number of feature maps and generalizes the feature maps by resolution reduction. Common pooling operations are that of taking the average and max of receptive cells over the input map \cite{Boureau10atheoretical}.

Due to the usage of convolutions for the extraction of partitioning features, \acp{CNN} require an assumption that the topology of the input dimensions provides some spatially regular sense of locality. Convolution on the regular grid is well documented and present in a variety of \ac{CNN} implementations \cite{vedaldi15matconvnet,jia2014caffe}, however when moving to domains that are not supported by the regular low-dimensional grid, convolution becomes an issue. Many application domains utilize irregular feature spaces \cite{Lowe1999ScaleInv}, and in such domains it may not be possible to define a spatial kernel filter or identify a method of translating such a kernel across spatial domain. Methods of handling such an irregular space as an input include using standard neural networks, embedding the feature space onto a grid to allow convolution \cite{Ijjina2014MOCAPCNN}, identifying local patches on the irregular manifold to perform geodesic convolutions \cite{MasciBBV15}, or graph signal processing based convolutions on graph signal data \cite{Henaff2015GraphCNN}. The potential applications of a convolutional network in the spatially irregular domain are expansive, however the graph convolution and pooling is not trivial, with graph representations of data being the topic of ongoing research \cite{Grady2010Graph,Zhang2015GSP}. The use of graph representation of data for deep learning is introduced by \cite{DBLP:journals/corr/BrunaZSL13}, utilizing the Laplacian spectrum for feature mining from the irregular domain. This is further expanded upon in \cite{Henaff2015GraphCNN}, providing derivative calculations for the backpropagation of errors during gradient descent. We formulate novel gradient equations that show more stable calculations in relation to both the input data and the tracked weights in the network.

In this methodology-focused study, we explore the use of graph based signal-processing techniques for convolutional networks on irregular domain problems. We evaluate two methods of graph pooling operators and report the effects of using interpolation in the spectral domain for identifying localized filters. We evaluate the use of Algebraic Multigrid agglomeration for graph pooling. We have also identified an alternative to the gradient calculations of \cite{Henaff2015GraphCNN} by formulating the gradients in regards to the input data as the spectral convolution of the gradients of the output with the filters (Equation \ref{eq:dataDer}), and the gradients for the weights as the spectral convolution of the input and output gradients (Equation \ref{eq:weightDer}). These proposed gradient calculations show consistent stability over previous methods \cite{Henaff2015GraphCNN}, which in turn benefits the gradient based training of the network. Results are reported on the MNIST dataset and the subsampled MNIST on an irregular grid.

The rest of the paper is outlined as follows. Section \ref{sec:meth} describes the generation of a graph based \ac{CNN} architecture, providing the convolution and pooling layers in the graph domain by use of signal-processing on the graph. Section \ref{sec:impl} details the experimental evaluation of the proposed methods and a comparison against the current state of the art, with Section \ref{sec:resu} reporting the results found and conclusions drawn in Section \ref{sec:conc}. \label{sec:intr}
\begin{figure*}
\begin{center}
\includegraphics[width=1\linewidth, trim= 0 500 0 0, clip=true]{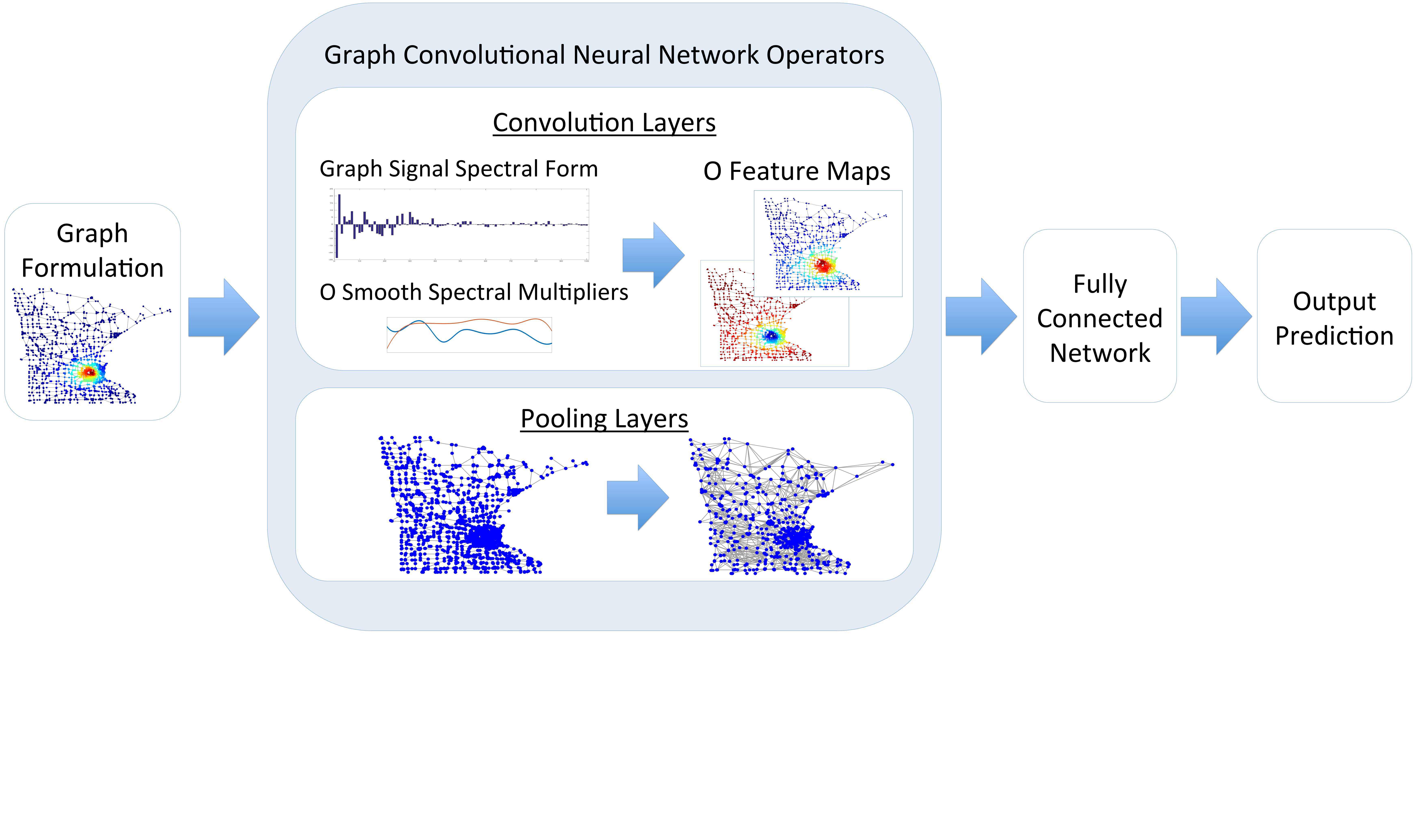}
\caption{Graph based Convolutional Neural Network components. The \ac{GCNN} is designed from an architecture of graph convolution and pooling operator layers. Convolution layers generate $O$ output feature maps dependent on the selected $O$ for that layer. Graph pooling layers will coarsen the current graph and graph signal based on the selected vertex reduction method.}
\label{fig:pipeline}
\end{center}
\end{figure*}
\section{Methods}\label{sec:meth}The familiar \ac{CNN} architecture pipeline consists of an input layer, a collection of convolution and/or pooling layers followed by a fully connected neural network and an output prediction layer. One issue with \acp{CNN} is that the convolution of a filter across the spatial domain is non-trivial when considering domains in which there is no regular structure. One solution is to utilize the multiplication in the spectral graph domain to perform convolution in the spatial domain, obtaining the feature maps via graph signal processing techniques. The graph based \ac{CNN} follows a similar architecture to standard \acp{CNN}; with randomly initialized spectral multiplier based convolution learnt in the spectral domain of the graph signal and graph coarsening based pooling layers, see Figure \ref{fig:pipeline} for a pipeline. Training is compromised of a feed-forward pass through the network to obtain outputs, with loss propagated backwards through the network to update the randomly initialized weights. 

The topic of utilizing graphs for the processing of signals is a recently emerging area in which the graph $G$ forms a carrier for the data signal $f$ \cite{Shuman2013SignalGraph}. The graph holds an underlying knowledge about the spatial relationship between the vertices and allows many common signal processing operators to be performed upon $f$ via $G$, such as wavelet filtering, convolution, and Fourier Transform \cite{Shuman2013SignalGraph,Hammond2011129}. By representing the observed domain as a graph it is possible to perform the signal processing operators on the observed data as graph signals. Coupling these graph signal processing techniques with deep learning it is possible to learn within irregularly spaced domains, upon which conventional \acp{CNN} would be unable to convolve a regular kernel across. The proposed technique will therefore open the door for deep learning to be utilized by a wider collection of machine learning and pattern recognition domains with irregular, yet spatially related features. 

\subsection{Convolution on Graph}
A graph $G=\{V,W\}$ consists of $N$ vertices $V$ and the weights $W$ of the undirected, non-negative, non-selflooping edges between two vertices $v_i$ and $v_j$. The unnormalized graph Laplacian matrix $L$ is defined as $L=D-W$, where $d_{i,i}=\sum_{i=1}^{N}{a_i}$ forms a diagonal matrix containing the sum of all adjacencies for a vertex. Given $G$, an observed data sample is a signal $f\in\mathbb{R}^N$ that resides on $G$, where $f_i$ corresponds to the signal amplitude at vertex $v_i$.

\begin{figure}[t!]
\begin{center}
\begin{minipage}[b]{0.4\linewidth}
\centerline{\adjincludegraphics[width=1\linewidth]{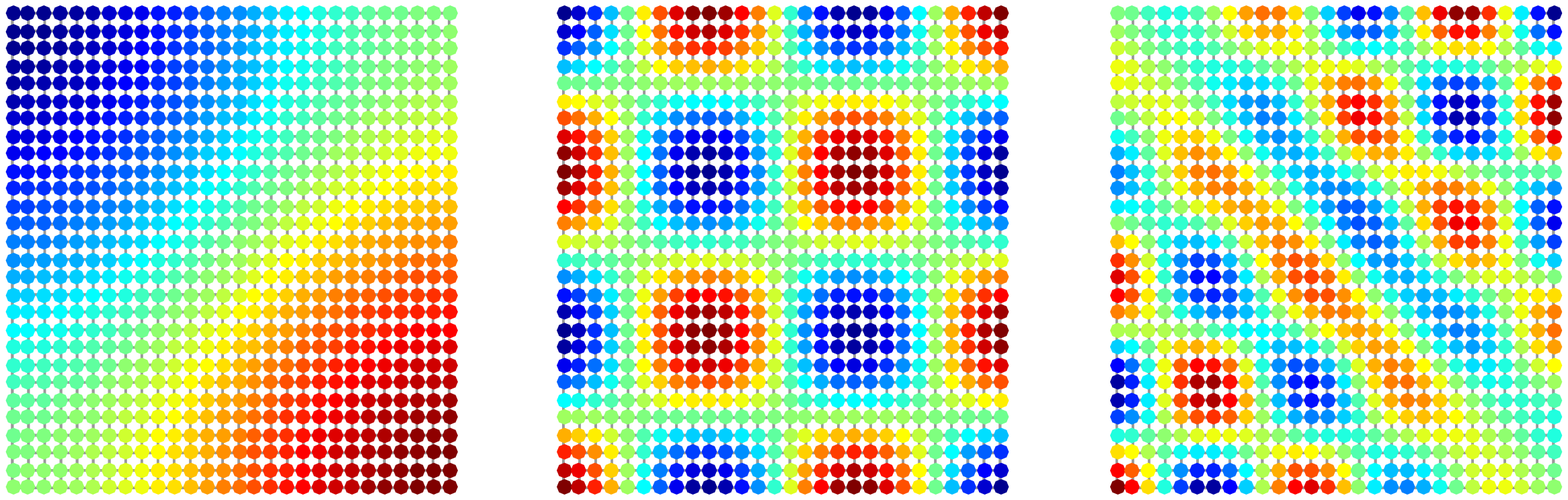}}
\end{minipage}
\hspace{1cm}
\begin{minipage}[b]{0.4\linewidth}
\centerline{\adjincludegraphics[width=1\linewidth]{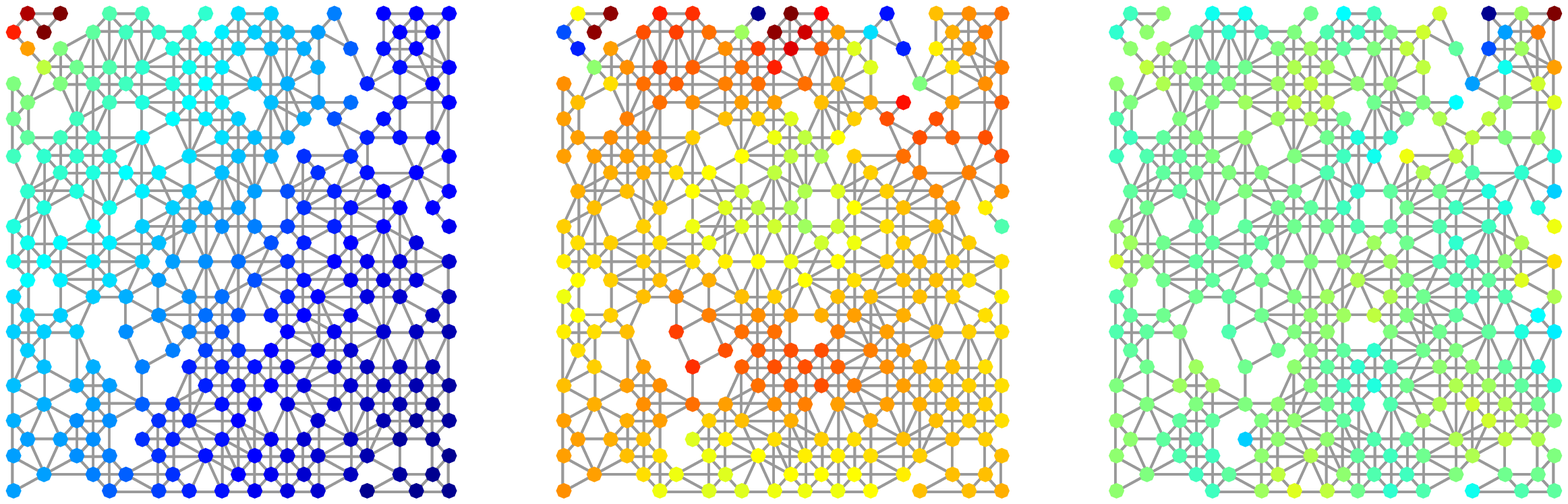}}
\end{minipage}
\caption{Eigenvectors $u_{i=\{2,20,40\}}$ of the full $28\times{}28$ regular gird (left) and the subsampled irregular grid (right).}
\label{fig:eigs_eps}
\end{center}
\end{figure}

Convolution is one of the two key operations in the \ac{CNN} architecture, allowing for locally receptive features to be highlighted in the input image \cite{Lecun1998CNN}. A similar operator is presented in graph based \ac{CNN}, however due to the potentially irregular domain graph convolution makes use of the convolution theorem of convolution in the spatial domain being a multiplication in the frequency domain \cite{Bracewell199ConvolutionTheorem}.\\
To project the graph signal into the frequency domain, the Laplacian $L$ is decomposed into a full matrix of orthonormal eigenvectors $U=\{u_{i=1\ldots{}N}\}$, where $u_i$ is a column of the matrix $U$, and the vector of associated eigenvalues $\lambda_{i=1\ldots{}N}$ \cite{Shuman2013SignalGraph}, Figure~\ref{fig:eigs_eps}. The forward Graph Fourier Transform is therefore given for a given signal as $\tilde{f}_{i}=\sum_{l=1}^{N}{\lambda_{l}f^{T}_{i}u_{i}}$, and its corresponding inverse $f_{i}=\sum_{l=1}^{N}{\lambda_{l}\tilde{f}_{i}u_{i}}$. Using the matrix $U$ the Fourier transform is defined as $\tilde{f}=U^{T}f$, and the inverse as $f=U\tilde{f}$, where $U^T$ is the transpose of the eigenvector matrix.

For forward convolution, a convolutional operator in the vertex domain can be composed as a multiplication in the Fourier space of the Laplacian operator \cite{Bracewell199ConvolutionTheorem}. Given the spectral form of our graph signal $\tilde{f}\in\mathbb{R}^N$ and the spectral multiplier $k\in{}\mathbb{R}^N$, the convolved output signal in the original spatial domain is the spectral multipication, i.e.\ $y = U\tilde{f}k$. It is possible to expand this for multiple input channels and multiple output feature maps: \begin{equation}
y_{s,o} = U\sum_{i=1}^{I}{U^{T}f_{s,i}\odot{}k_{i,o}}\ ,
\end{equation} where $I$ is the number of input channels for $f$, $s$ is a given batch sample, and $o$ indexes an output feature map from $O$ output maps. 

Localized regions in the spatial domain are defined by the kernel receptive field in \acp{CNN}, and for graph based \acp{CNN} the spatial vertex domain localization is given by a smoothness within the spectral domain. Therefore to identify local features within the spatial domain the spectral multipliers used for spectral convolution are identified by tracking a subsampled set of filter weights $\hat{k}_{i,o}k\in{}\mathbb{R}^{<N}$ which are interpolated up to a full filter via a smoothing kernel $\Phi$ such as cubic splines: $k_{i,o} = \Phi\hat{k}_{i,o}$. This has the added benefit of reducing the number of tracked weights, however leads to an extra pair of operations in interpolating the weights to the full $k\in{}\mathbb{R}^N$ for multiplication. Reducing the number of tracked weights increases the smoothness of the final interpolated filter, and lowering the tuning parameter of the number of tracked weights learns sharper features.


\subsection{Backpropagation on Graph}
Backpropagation of errors is a pivotal component of deep learning, providing updates of weights and bias for the networks towards the target function with gradient descent. This requires obtaining derivatives in regards to the input and weights used to generate the output, in the case of graph based \ac{CNN} convolution the gradients are formulated in regards to the graph signal $f$ and the spectral multipliers $k$. The gradients for an input feature map channel $f_{s,i}$ is given as the convolution of the gradients for the output $\nabla{}y$ and the spectral multipliers in the spectral domain via
\begin{equation}\label{eq:dataDer}
\nabla{}f_{s,i} = U\sum_{o=1}^{O}{U^{T}\nabla{}y_{s,o}\odot{}k_{i,o}}
\end{equation}
for a provided batch of $S$ graph signals. Gradients for the full set of interpolated spectral multipliers is formulated as the convolution of the gradients for the output $\nabla{}y$ with the input $f_{s,i}$ via
\begin{equation}\label{eq:weightDer}
\nabla{}k_{i,o} = \sum_{s=1}^{N}{U^{T}\nabla{}y_{s,o}\odot{}U^{T}f_{s,i}}.
\end{equation}
As the filters are spectral domain multipliers, we do not project this spectral convolution back through the graph Fourier transform. The smooth multiplier weights $\nabla{}k$ can then be projected back to the subsampled set of tracked weights by the multiplication with the inversed smoothing kernel $\nabla{}\hat{k}_{i,o} = \Phi^{T}\nabla{}k_{i,o}$.

\subsection{Pooling on Graph}
The pooling layer is the second component in conventional \ac{CNN}, reducing the resolution of the input feature map in both an attempt to generalize the features identified and to manage the memory complexity when using numerous filters \cite{Boureau10atheoretical}. During graph based convolutions there is no reduction in size between the input signal and the output feature map due to the multiplication of the $\mathbb{R}^N$ filter with the $\mathbb{R}^N$ spectral signal. As such, each layer of a deep graph \ac{CNN} would possess a graph with $\mathbb{R}^N$ vertices. Such a construction could be beneficial, as this would allow the algorithm to store a single instance of the graph and the associated $N^2$ eigenvector matrix $U$. If pooling is utilized, there is benefit gained from the feature map generalization and the reduction in complexity of the graph Fourier transforms as each layer's vertex count $N$ is lowered. To pool local features together on the graph, it is required to perform graph coarsening and project the input feature maps through to the new, reduced size graph. Coarsening $G = \{V,W\}$ to $\hat{G}=\{\hat{V},\hat{W}\}$ not only requires the reduction of vertex counts, but also a handling of edges between the remaining $\hat{N}$ vertices. Common methods of generating $\hat{V}$ are to either select a subset of $V$ to carry forward to $\hat{G}$ \cite{Shuman2016Kron} or to form completely new set of nodes $\hat{V}$ from some aggregation of related nodes within $V$ \cite{Safro09comparisonof}.\\
One method for selection of $\hat{V}$ can be achieved by selecting the largest eigenvalue $\lambda_N$ and splitting $V$ into two subsets based on the polarity of the associated eigenvector $U_N$ \cite{Liu2014GSPCoarsen}. We can therefore define $\hat{V} = \{u_N,i\}; u_N,i >= 0$ and its complement $\hat{V}^c = \{u_N,i\}; u_N,i < 0$. $\hat{V}$ is then utilized to construct the graph $\hat{G}$, although by reversing the selection for the polarity to keep it is just as understandable to choose $\hat{V}^c$ for construction of $\hat{G}$. 

In this study we utilize \ac{AMG} for graph coarsening, a method of projecting signals to a coarser graph representation obtained via greedy selection of vertices \cite{Safro09comparisonof}. Aggregation takes a subset of vertices on $V$ and generates a singular vertex in the new set of coarsened nodes $\hat{V}$ in the output graph. Graph coarsening is by no means a trivial task, with extensive literature exploring the subject \cite{Liu2014GSPCoarsen,Safro2012GraphCoarsening,Safro09comparisonof}.

\begin{figure}[t!]
\begin{center}
\begin{minipage}[b]{0.4\linewidth}
\centerline{\adjincludegraphics[width=1\linewidth , trim={0 0 0 {.5\height}},clip=true]{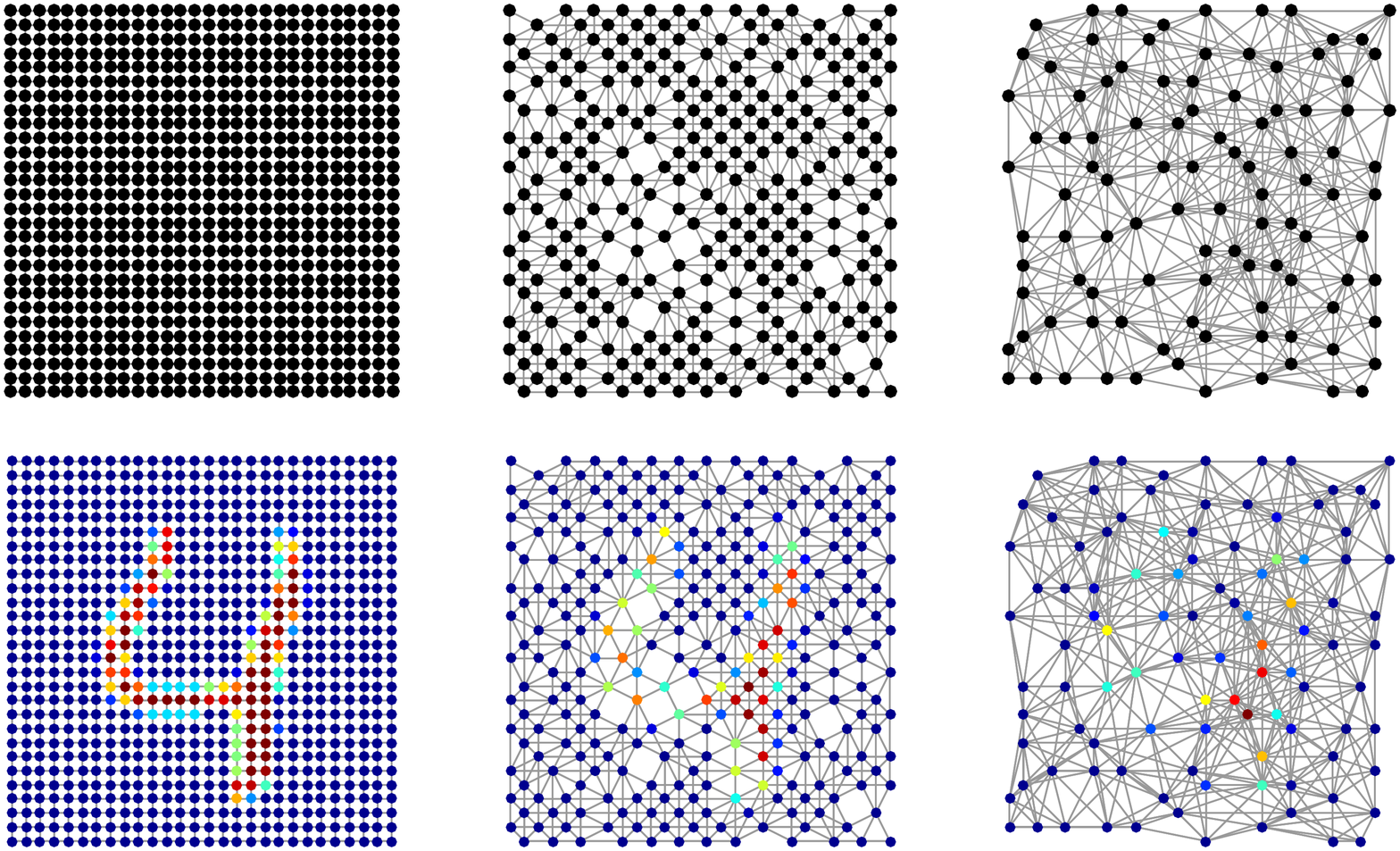}}
\end{minipage}
\hspace{1cm}
\begin{minipage}[b]{0.4\linewidth}
\centerline{\adjincludegraphics[width=1\linewidth , trim={0 0 0 {.46\height}},clip=true]{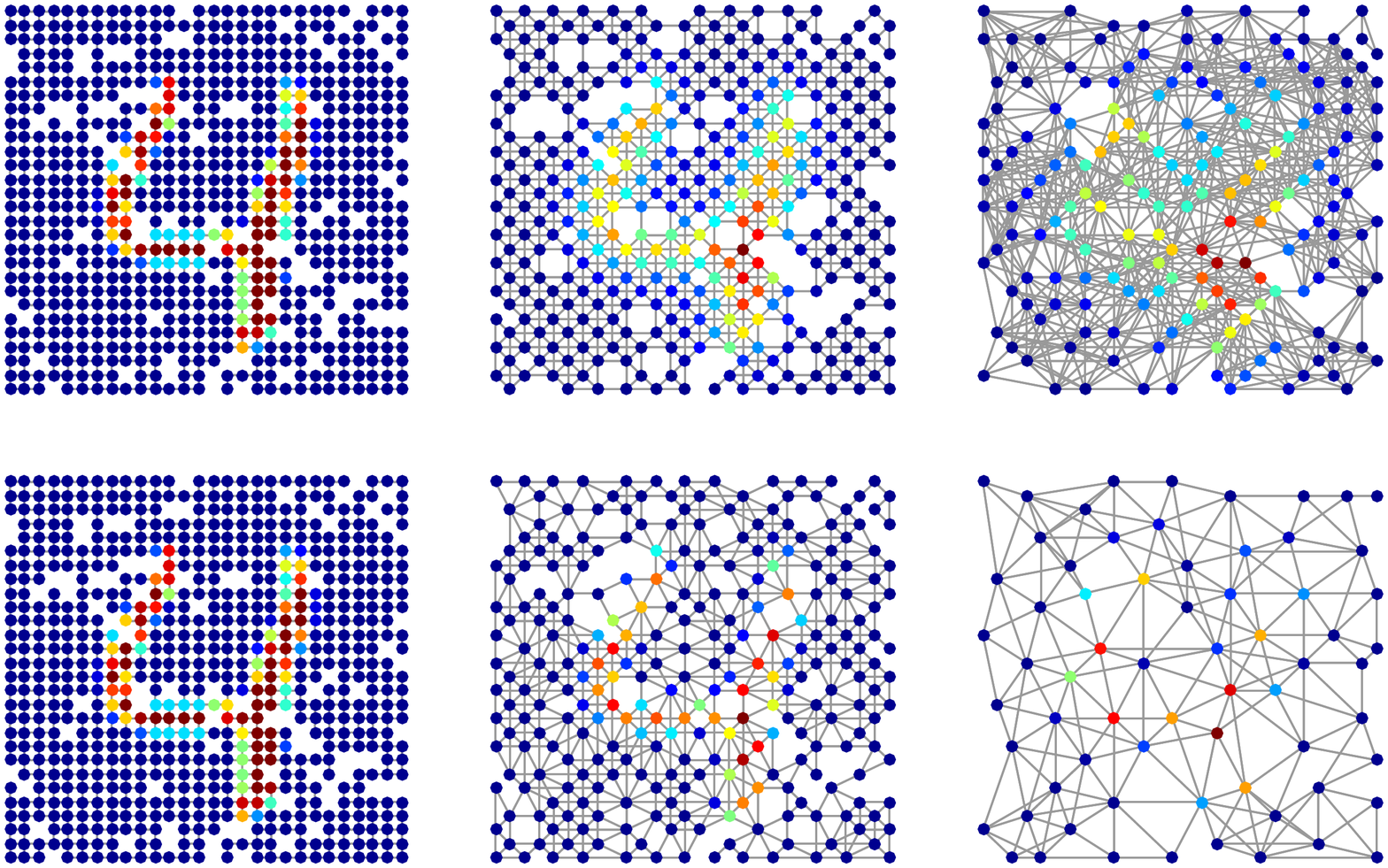}}
\end{minipage}
\caption{Two levels of graph pooling operation on regular and irregular grid with MNIST signal. From left: Regular grid, \ac{AMG} level 1, \ac{AMG} level 2, Irregular grid, \ac{AMG} level 1, \ac{AMG} level 2.}
\label{fig:subpool_eps}
\end{center}
\end{figure}

%

With a coarser graph structure $\hat{G}$ it is required to then downsample the graph signal $f_{1:N}$ into a new signal $\hat{f}_{1:n}$ that is able to reside on $\hat{G}$. \ac{AMG} provides a set of matrices for the interpolation of the input signal $f$; the restriction matrix $R$ and the projection matrix $P$. Downsampling $f\in\mathbb{R}^N$ on $G$ to $\hat{f}\in\mathbb{R}^{\hat{N}}$ on $\hat{G}$ is achieved by the multiplication of the signal with the restriction matrix, $\hat{f}_{s,i}=Rf_{s,i}$, whilst the reverse pooling required for backpropagation is achieved via multiplication with the projection matrix, $f_{s,i}=P\hat{f}_{s,i}$.



\section{Implementation} \label{sec:impl}
Although we utilize forms of the 2D grid, the graph CNN is generalizable to more irregular domain problems; such as sensor networks, mesh signals, text corpora, human skeleton graphs and more. These domains quite often contain irregular spatial geometries, upon which it is non-trivial to define a filter kernel for convolution. In this study we evaluate the performance of the proposed graph CNN derivative calculations with an implementation on both the standard regular 28 grid and the irregular subsampled 2D grid. 

The 2D grid is the graph representation of the Von Neumann neighborhood of vertices in a regular domain, most commonly applied to that of pixel relationships in images. For an image, each pixel is represented by a vertex on $G$, with the pixel intensities for each vertex forming the graph signal $f$. The edge weights are taken as the euclidean distance between the nodes in the Von Neumann neighborhood. To evaluate the performance of graph \ac{CNN} on the 2D grid we utilize the MNIST dataset, consisting of 60,000 examples of handwritten numerical digits in $28\times{}28$ grayscale pixel images. The edge weights for $G$ are the binary presence of an edge between $v_i$ and $v_j$ on the 4-way adjacency, with $V\in\mathbb{R}^{784}$.

To obtain an irregular spatial geometry domain upon which a conventional \ac{CNN} cannot convolve, we subsampled the $28\times{}28$ grid by selecting 84 random vertices to exclude from the grid. Upon removing the selected vertices and their corresponding edges from the graph, we then subsample the MNIST dataset with the respective signals such that $f\in\mathbb{R}^{700}$. This irregular spatial domain now requires the graph-based \ac{CNN} operators above to form a convolved output feature map.

The architecture of the graph \ac{CNN} was set to $C^{20}PC^{50}PRF$; where $C^{\kappa{}}$ defines a convolutional layer with 60 tracked weights and $\kappa{}$ output feature maps, P defines an \ac{AMG} pooling with a coarsening factor of $\beta{}=0.05$ and 2 levels, $R$ defines a rectified linear unit layer, and finally $F$ describes fully connected layers providing output class predictions. Networks were trained for 500 epochs, with the full 10,000 test samples being classified at each epoch to track the predictive performance of the network.

To perform derivative checking, the calculation of the gradients for $\nabla{}f$, $\nabla{}k$ and $\nabla{}\hat{k}$ were evaluated using random perturbations of errors on the scale of $10^{-4}$. Derivatives for $\nabla{}\hat{k}$ were checked for interpolation over varying numbers of tracked weights in the network, including the full set $\hat{N}\in\mathbb{R}^N$. The experiment was repeated 100 times and the average percentage error of the calculated gradient versus the empirically obtained gradient is reported in Figure \ref{fig:gradErr}.

The graph-based CNN architecture was implemented within MATLAB using GPU enabled operators.
\section{Results} \label{sec:resu}The graph \ac{CNN} method was evaluated on both the regular $28\times28$ grid and irregular randomly subsampled grid. We report the predictive accuracy of the network at each epoch of training using both the proposed graph \ac{CNN} method and the method proposed by \cite{Henaff2015GraphCNN}. We also show the effects of smoothed spectral multiplier filters on the convolution output and the derivative errors we obtained for gradient calculations. In summary we found that by increasing the smoothness of the spectra filters we were able to increase the local relationship of features in the spatial domain, however this also resulted in higher error being introduced by the interpolation when calculating the gradients of the tracked weights $\hat{k}$. Overall we found that the proposed calculations for derivatives in respect to $\hat{k}$ introduced little error during backpropagation. The accuracy observed when testing unobserved samples is very promising, exceeding 94\% on both the regular and irregular geometry domains.

\subsection{Convolution and filter smoothness}
\begin{figure}[t!]
\begin{center}
\centerline{\includegraphics[width=1\linewidth, clip=true]{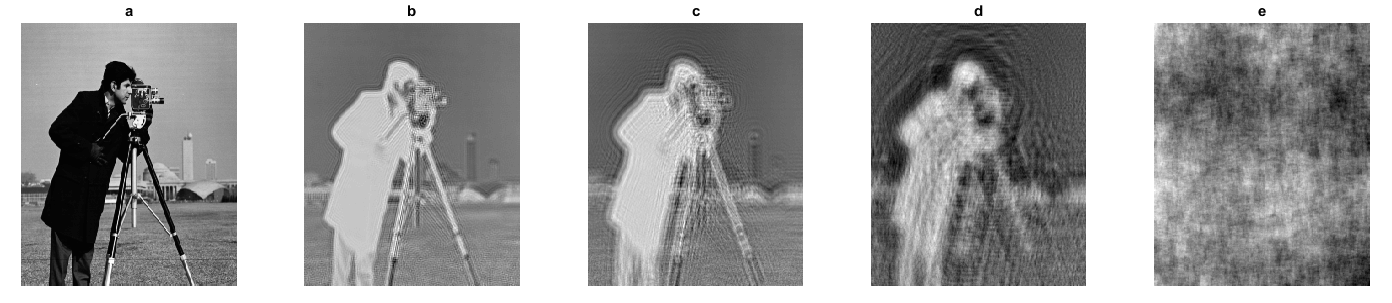}}
\caption{Effect of spline interpolation of tracked weights on spectral filter smoothness during graph-based convolution. a) Original image, b) $\hat{k}=\mathbb{R}^{\mathrm{ceil}(\sqrt[\leftroot{-2}\uproot{2}4]{N})}$, c) $\hat{k}=\mathbb{R}^{\mathrm{ceil}(\sqrt[\leftroot{-2}\uproot{2}3]{N})}$, d) $\hat{k}=\mathbb{R}^{\mathrm{ceil}(\sqrt[\leftroot{-2}\uproot{2}2]{N})}$, e) $\hat{k}=\mathbb{R}^N$}
\label{fig:smoothFilters}
\end{center}
\end{figure}

Reducing the number of tracked filter weights produces smoother spectral multipliers after interpolation up to $k_{i,o} = \Phi\hat{k}_{i,o}$. Figure~\ref{fig:smoothFilters} shows the effect of interpolating weights from various lengths of $\hat{k}$ as applied to the 2D graph with the Cameraman\bmvaOneDot{}tif graph signal residing on it. As the number of tracked weights is reduced the spatial locality of the features learnt is reduced, providing sharper features, whilst as the number of tracked weights approaches $N$ the spatial localization of the feature map is lost. 
 
\subsection{Localized feature maps}
\begin{figure}[t!]
\begin{center}
\centerline{\adjincludegraphics[width=0.90\linewidth, clip=true]{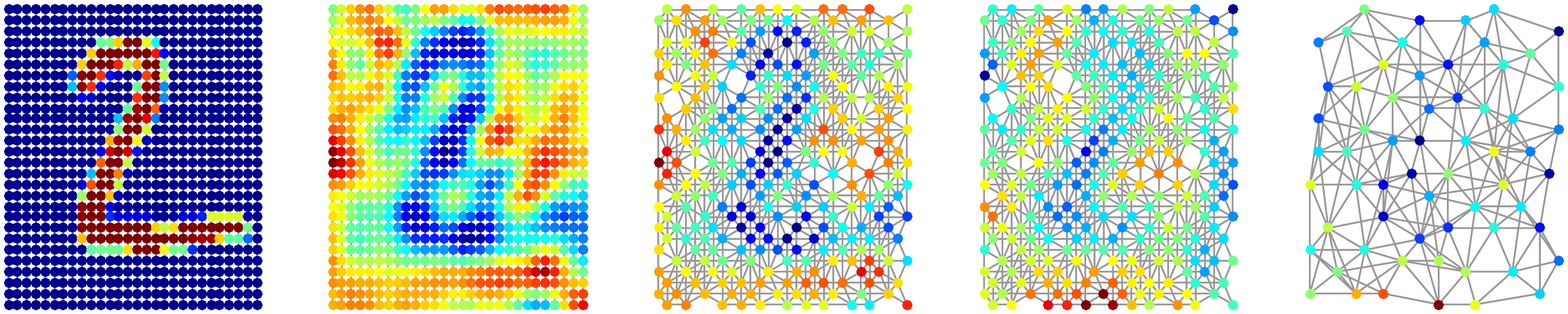}}
\caption{Feature maps formed by a feed-forward pass of the regular domain. From left: Original image, Convolution round 1, Pooling round 1, Convolution round 2, Pooling round 2.}
\centerline{\adjincludegraphics[width=0.90\linewidth, clip=true]{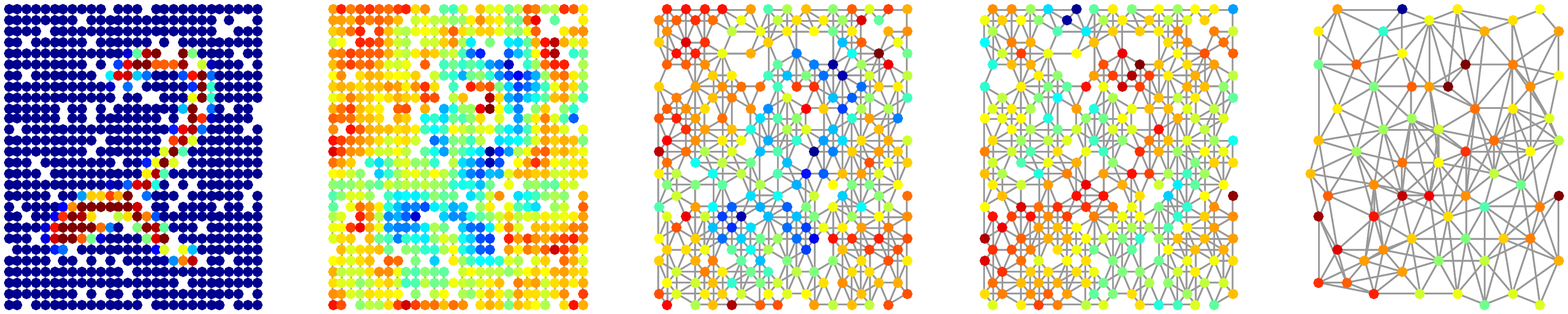}}
\caption{Feature maps formed by a feed-forward pass of the irregular domain. From left: Original image, Convolution round 1, Pooling round 1, Convolution round 2, Pooling round 2.}
\label{fig:resMaps}
\end{center}
\vspace{-0.2in}
\end{figure}

By interpolating smooth spectral multipliers from the 60 tracked weights we were able to convolve over the irregular domain to produce feature maps in the spatial domain with spatially localized features. Figure \ref{fig:resMaps} visualizes output for each layer of the Graph CNN convolution and pooling layers for both the regular and irregular domain graphs. 

\subsection{Backpropagation derivative checks}
\begin{figure}[t!]
\begin{center}
\centerline{\adjincludegraphics[width=0.9\linewidth, trim= 100 0 100 0, clip=true]{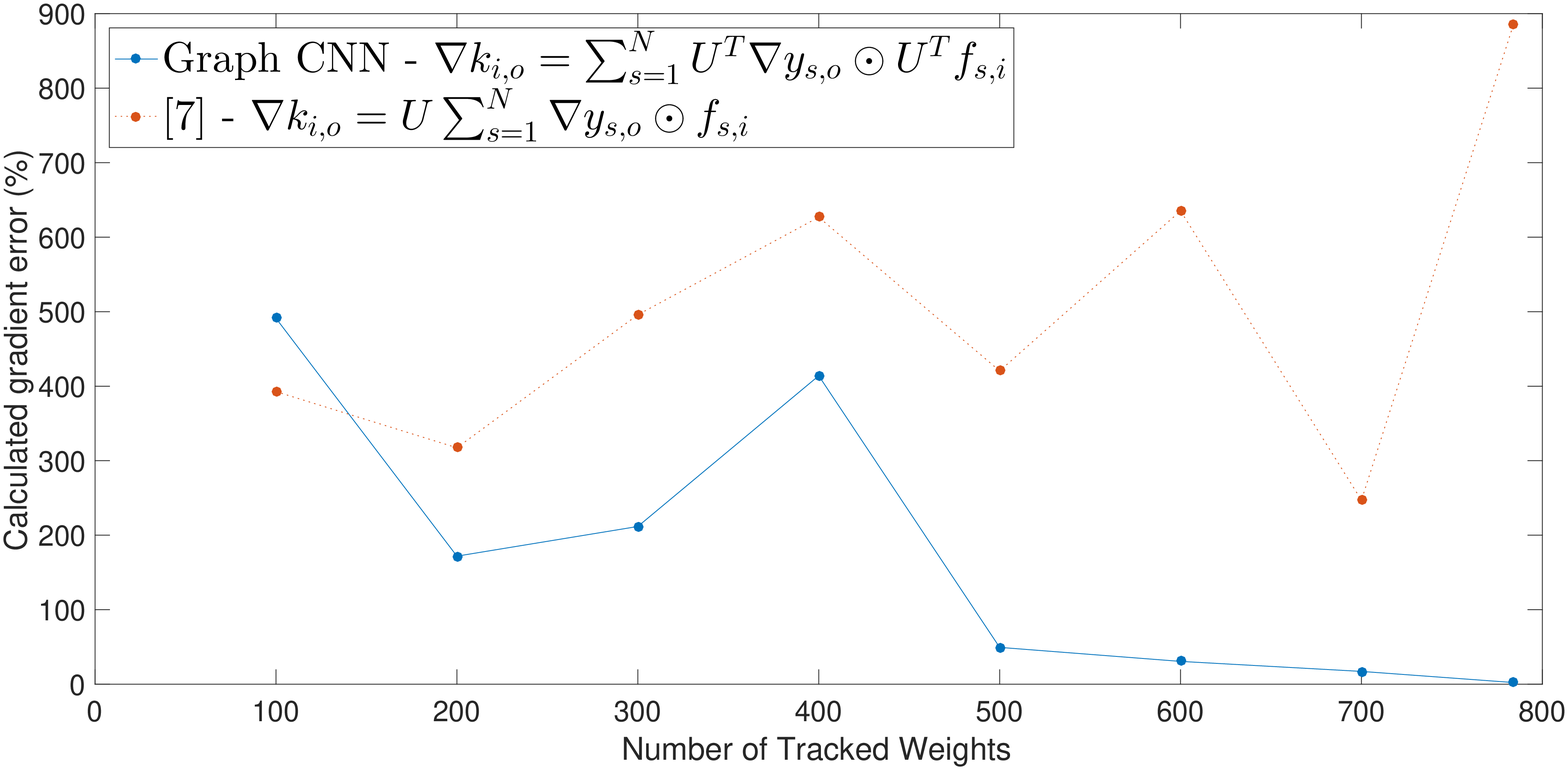}}
\caption{Gradient calculation errors for interpolation of various numbers of tracked weights.}
\label{fig:gradErr}
\vspace{-0.2in}
\end{center}
\end{figure}

The proposed method gave an average of $1.41\%(\pm 4.00\%)$ error in the calculation of the gradients for the input feature map. In comparison, by not first applying a graph Fourier transform to $\nabla{}y_{s,o}$ in the calculation for $\nabla{}f_{s,i}$, as in \cite{Henaff2015GraphCNN}, we obtain errors of $376.50\% (\pm1020.79\%)$. Similarly the proposed method of obtaining the spectral forms of $\nabla{}y_{s,o}$ and $f_{s,i}$ in the calculation of $\nabla{}k_{i,o}$ gave errors of $3.81\%(\pm{}16.11\%)$. By not projecting to the spectral forms of these inputs, errors of $826.08\%(\pm{}4153.32\%)$ are obtained. Figure~\ref{fig:gradErr} shows the average percentage derivative calculation error for $\nabla{}\hat{k}$ of varying numbers of tracked weights over 100 runs. The proposed method of gradient calculation shows lower errors than the compared method gradient calculation of $\nabla{}k$ when $\hat{k}\in\mathbb{R}^N$ and all but the lowest number of tracked weights of $\hat{k}\in\mathbb{R}^{100}$. The introduction of interpolation leads to a higher introduction of error into the calculated gradient errors, especially in the presence of a low number of tracked weights. 


\subsection{Testing performance}
\begin{figure}[t!]
\begin{center}
\centerline{\adjincludegraphics[width=0.9\linewidth, trim= 100 0 100 0, clip=true]{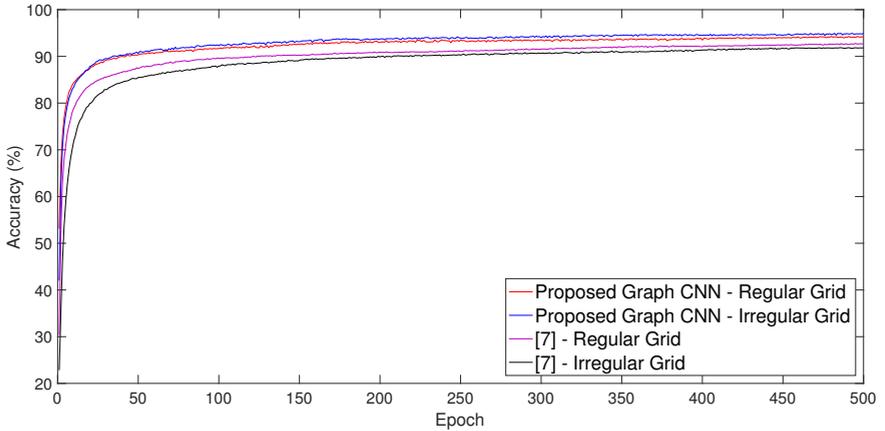}}
\caption{Test Set accuracy on the MNIST dataset on the regular and irregular 2D grid. An increasing in testing accuracy is observed when utilizing the proposed gradient calculations from equations \ref{eq:dataDer} and \ref{eq:weightDer}.}
\label{fig:res}
\end{center}
\vspace{-0.4in}
\end{figure}

Classification performance on the MNIST dataset is reported in Table~\ref{tbl:results}, with progression of testing accuracy over epochs given in Figure~\ref{fig:res} comparing between the proposed gradient calculations and those of \cite{Henaff2015GraphCNN}. The proposed graph CNN method does not obtain the 99.77\% accuracy rates of the state of the art \ac{CNN} architecture presented by \cite{Ciresan2012MCDNN} on the full \mbox{$28\times{}28$} grid. This is understandable, as standard \acp{CNN} are designed to operate in the regular Cartesian space, giving it a strong performance in the image classification problem. The main benefit of the graph \ac{CNN} is in it's ability to handle the irregular spatial domain presented by the subsampled MNIST grid by use of convolution in the graph spectral domain.

\begin{table}
\caption{Testing set accuracy of network (\%)}\label{tbl:results}
\begin{center}
\begin{tabular}{|c|c|c|}
\hline
Dataset  & \cite{Henaff2015GraphCNN} & Proposed Graph CNN\\\hline
Regular grid MNIST  & 92.69 & \bf{94.23}\\
Subsampled irregular grid MNIST & 91.84 & \bf{94.96} \\\hline
\end{tabular}
\end{center}
\end{table}



\section{Conclusion} \label{sec:conc}
This study proposes a novel method of performing deep convolutional learning on the irregular graph by coupling standard graph signal processing techniques and backpropagation based neural network design. Convolutions are performed in the spectral domain of the graph Laplacian and allow for the learning of spatially localized features whilst handling the non-trivial irregular kernel design. Results are provided on both a regular and irregular domain classification problem and show the ability to learn localized feature maps across multiple layers of a network. A graph pooling method is provided that agglomerates vertices in the spatial domain to reduce complexity and generalize the features learnt. GPU performance of the algorithm improves upon training and testing speed, however further optimization is needed. Although the results on the regular grid are outperformed by standard \ac{CNN} architecture this is understandable due to the direct use of a local kernel in the spatial domain. The major contribution over standard \acp{CNN} is the ability to function on the irregular graph is not to be underestimated. Graph based \ac{CNN} requires costly forward and inverse graph Fourier transforms, and this requires some work to enhance usability in the community. Ongoing study into graph construction and reduction techniques is required to encourage uptake by a wider range of problem domains.

\bibliography{references_acronyms,references}

\begin{thebibliography}{22}
\providecommand{\natexlab}[1]{#1}
\providecommand{\url}[1]{\texttt{#1}}
\expandafter\ifx\csname urlstyle\endcsname\relax
  \providecommand{\doi}[1]{doi: #1}\else
  \providecommand{\doi}{doi: \begingroup \urlstyle{rm}\Url}\fi

\bibitem[Boureau et~al.(2010)Boureau, Ponce, and Lecun]{Boureau10atheoretical}
Y-Lan Boureau, Jean Ponce, and Yann Lecun.
\newblock A theoretical analysis of feature pooling in visual recognition.
\newblock In \emph{Proc. Int. Conf. Mach. Learning}, 2010.

\bibitem[Bracewell(1999)]{Bracewell199ConvolutionTheorem}
Ronald Bracewell.
\newblock \emph{{The Fourier Transform \& Its Applications}}.
\newblock McGraw, 1999.

\bibitem[Bruna et~al.(2013)Bruna, Zaremba, Szlam, and
  LeCun]{DBLP:journals/corr/BrunaZSL13}
Joan Bruna, Wojciech Zaremba, Arthur Szlam, and Yann LeCun.
\newblock Spectral networks and locally connected networks on graphs.
\newblock \emph{CoRR}, abs/1312.6203, 2013.

\bibitem[Ciresan et~al.(2012)Ciresan, Meier, and Schmidhuber]{Ciresan2012MCDNN}
Dan~C. Ciresan, Ueli Meier, and J{\"{u}}rgen Schmidhuber.
\newblock Multi-column deep neural networks for image classification.
\newblock \emph{CoRR}, abs/1202.2745, 2012.

\bibitem[Grady and Polimeni(2010)]{Grady2010Graph}
Leo Grady and Jonathan~R. Polimeni.
\newblock \emph{Discrete Calculus - Applied Analysis on Graphs for
  Computational Science.}
\newblock Springer, 2010.

\bibitem[Hammond et~al.(2011)Hammond, Vandergheynst, and
  Gribonval]{Hammond2011129}
David~K. Hammond, Pierre Vandergheynst, and R{\'{e}}mi Gribonval.
\newblock Wavelets on graphs via spectral graph theory.
\newblock \emph{Applied and Computational Harmonic Analysis}, 30\penalty0
  (2):\penalty0 129 -- 150, 2011.

\bibitem[Henaff et~al.(2015)Henaff, Bruna, and LeCun]{Henaff2015GraphCNN}
Mikael Henaff, Joan Bruna, and Yann LeCun.
\newblock Deep convolutional networks on graph-structured data.
\newblock \emph{CoRR}, abs/1506.05163, 2015.

\bibitem[Ijjina and Mohan(2014)]{Ijjina2014MOCAPCNN}
E.~P. Ijjina and C.~K. Mohan.
\newblock Human action recognition based on mocap information using convolution
  neural networks.
\newblock In \emph{Proc. Int. Conf. Mach. Learning and Applications}, pages
  159--164, Dec 2014.
\newblock \doi{10.1109/ICMLA.2014.30}.

\bibitem[Jia et~al.(2014)Jia, Shelhamer, Donahue, Karayev, Long, Girshick,
  Guadarrama, and Darrell]{jia2014caffe}
Yangqing Jia, Evan Shelhamer, Jeff Donahue, Sergey Karayev, Jonathan Long, Ross
  Girshick, Sergio Guadarrama, and Trevor Darrell.
\newblock Caffe: Convolutional architecture for fast feature embedding.
\newblock \emph{arXiv:1408.5093}, 2014.

\bibitem[Lecun et~al.(1998)Lecun, Bottou, Bengio, and Haffner]{Lecun1998CNN}
Y.~Lecun, L.~Bottou, Y.~Bengio, and P.~Haffner.
\newblock Gradient-based learning applied to document recognition.
\newblock \emph{Proceedings of the IEEE}, 86\penalty0 (11):\penalty0
  2278--2324, 1998.

\bibitem[Li et~al.(2015)Li, Lin, Shen, Brandt, and Hua]{Li2015CNNCascade}
H.~Li, Z.~Lin, X.~Shen, J.~Brandt, and G.~Hua.
\newblock A convolutional neural network cascade for face detection.
\newblock In \emph{Proc. IEEE Conf. on Comp. Vis. and Pat. Rec.}, pages
  5325--5334, 2015.

\bibitem[Liu et~al.(2014)Liu, Wang, and Gu]{Liu2014GSPCoarsen}
P.~Liu, X.~Wang, and Y.~Gu.
\newblock Graph signal coarsening: Dimensionality reduction in irregular
  domain.
\newblock In \emph{IEEE GLobal Conf. Signal and Information Processing}, pages
  798--802, 2014.

\bibitem[Lowe(1999)]{Lowe1999ScaleInv}
D.~G. Lowe.
\newblock Object recognition from local scale-invariant features.
\newblock In \emph{Proc. Int. Conf. on Comp. Vis.}, volume~2, pages 1150--1157,
  1999.

\bibitem[Masci et~al.(2015)Masci, Boscaini, Bronstein, and
  Vandergheynst]{MasciBBV15}
Jonathan Masci, Davide Boscaini, Michael~M. Bronstein, and Pierre
  Vandergheynst.
\newblock Shapenet: Convolutional neural networks on non-euclidean manifolds.
\newblock \emph{CoRR}, abs/1501.06297, 2015.

\bibitem[Oquab et~al.(2014)Oquab, Bottou, Laptev, and
  Sivic]{Oquab2014CNNTransfer}
M.~Oquab, L.~Bottou, I.~Laptev, and J.~Sivic.
\newblock Learning and transferring mid-level image representations using
  convolutional neural networks.
\newblock In \emph{Proc. IEEE Conf. on Comp. Vis. and Pat. Rec.}, pages
  1717--1724, 2014.

\bibitem[Safro(2009)]{Safro09comparisonof}
Ilya Safro.
\newblock Comparison of coarsening schemes for multilevel graph partitioning.
\newblock In \emph{Int. Conf. Learning and Intelligent Optimization}, pages
  191--205. Springer-Verlag, 2009.

\bibitem[Safro et~al.(2012)Safro, Sanders, and
  Schulz]{Safro2012GraphCoarsening}
Ilya Safro, Peter Sanders, and Christian Schulz.
\newblock \emph{Proc. Int. Symposium Experimental Algorithms}, chapter Advanced
  Coarsening Schemes for Graph Partitioning, pages 369--380.
\newblock Springer Berlin Heidelberg, 2012.

\bibitem[Shuman et~al.(2016)Shuman, Faraji, and Vandergheynst]{Shuman2016Kron}
D.~I. Shuman, M.~J. Faraji, and P.~Vandergheynst.
\newblock A multiscale pyramid transform for graph signals.
\newblock \emph{IEEE Trans. Signal Process.}, 64\penalty0 (8):\penalty0
  2119--2134, 2016.

\bibitem[Shuman et~al.(2013)Shuman, Narang, Frossard, Ortega, and
  Vandergheynst]{Shuman2013SignalGraph}
D.I. Shuman, S.K. Narang, P.~Frossard, A.~Ortega, and P.~Vandergheynst.
\newblock The emerging field of signal processing on graphs: Extending
  high-dimensional data analysis to networks and other irregular domains.
\newblock \emph{IEEE Signal Processing Magazine}, 30\penalty0 (3):\penalty0
  83--98, 2013.

\bibitem[Vedaldi and Lenc(2015)]{vedaldi15matconvnet}
A.~Vedaldi and K.~Lenc.
\newblock Matconvnet -- convolutional neural networks for matlab.
\newblock In \emph{Proceeding of the {ACM} Int. Conf. on Multimedia}, 2015.

\bibitem[Zhang et~al.(2015)Zhang, Florencio, and Chou]{Zhang2015GSP}
Cha Zhang, Dinei Florencio, and Philip Chou.
\newblock Graph signal processing - a probabilistic framework.
\newblock Technical Report MSR-TR-2015-31, 2015.

\bibitem[Zhang and Zhou(2006)]{Zhang2006NNGenomes}
Min-Ling Zhang and Zhi-Hua Zhou.
\newblock Multilabel neural networks with applications to functional genomics
  and text categorization.
\newblock \emph{IEEE Transactions on Knowledge and Data Engineering},
  18\penalty0 (10):\penalty0 1338--1351, 2006.

\end{thebibliography}
\end{document}